%% file: main.tex
\documentclass[11pt,a4paper]{article}
\usepackage[hyperref]{eacl2021}
\usepackage{times}
\usepackage{latexsym}
\usepackage{graphicx}
\usepackage{amsmath}
\usepackage{flexisym}
\usepackage{multirow}
\usepackage{array}
\usepackage{float}
\usepackage{dblfloatfix}
\usepackage{footmisc}

\renewcommand*{\thefootnote}{\fnsymbol{footnote}}

\usepackage{microtype}

\aclfinalcopy 
\def\aclpaperid{194} 

\title{MONAH: Multi-Modal Narratives for Humans to analyze conversations}

\author{
\begin{tabular}{>{\centering\arraybackslash}m{4.9cm}
                  >{\centering\arraybackslash}m{5.2cm}
                  >{\centering\arraybackslash}m{4.9cm}
                }
Joshua Y. Kim  & Greyson Y. Kim & Chunfeng Liu\\
\normalfont University of Sydney 
& \normalfont Success Beyond Pain
& \normalfont Hello Sunday Morning\\

\normalfont New South Wales, Australia & \normalfont \normalfont Western Australia, Australia &
\normalfont New South Wales, Australia \\

\\
Rafael A. Calvo & Silas C.R. Taylor & Kalina Yacef\footnotemark \\
\normalfont Imperial College London & \normalfont University of New South Wales & \normalfont University of Sydney\\
\normalfont London, United Kingdom &
\normalfont New South Wales, Australia &  \normalfont New South Wales, Australia\\

\end{tabular}
}

\date{}

\begin{document}
\maketitle

\footnotetext{* Corresponding author (kalina.yacef@sydney.edu.au)}

\setcounter{footnote}{0} 
\renewcommand*{\thefootnote}{\arabic{footnote}}

\begin{abstract}
In conversational analyses, humans manually weave multimodal information into the transcripts, which is significantly time-consuming. We introduce a system that automatically expands the verbatim transcripts of video-recorded conversations using multimodal data streams. This system uses a set of preprocessing rules to weave multimodal annotations into the verbatim transcripts and promote interpretability. Our feature engineering contributions are two-fold: firstly, we identify the range of multimodal features relevant to detect rapport-building; secondly, we expand the range of multimodal annotations and show that the expansion leads to statistically significant improvements in detecting rapport-building.
\end{abstract}

\section{Introduction}
\label{sect:introduction}

Dyadic human-human dialogs are rich in multimodal information. Both the visual and the audio characteristics of how the words are said reveal the emotions and attitudes of the speaker. Given the richness of multimodal information, analyzing conversations requires both domain knowledge and time. The discipline of conversational analysis is a mature field. In this discipline, conversations could be manually transcribed using a technical system developed by \citet{jefferson2004glossary}, containing information about intonation, lengths of pauses, and gaps. Hence, it captures both \textit{what} was said and \textit{how} it was said\footnote{Please visit www.universitytranscriptions.co.uk/jefferson-transcription-example/ for an audio example.}. However, such manual annotations take a great deal of time. Individuals must watch the conversations attentively, often replaying the conversations to ensure completeness.

Automated \citet{jefferson2004glossary} transcripts could be generated from video-recordings \cite{moore2015automated}. However, the potential issue with Jeffersonian annotations is that there are often within-word annotations and symbols which makes it hard to benefit from pre-trained word embeddings. Inspired by the Jeffersonian annotations, we expand the verbatim transcripts with multimodal annotations such that downstream classification models can easily benefit from pre-trained word embeddings.

Our paper focuses on the classification task of predicting rapport building in conversations. Rapport has been defined as a state experienced in interaction with another with interest, positivity, and balance \cite{cappella1990defining}. If we can model rapport building in the medical school setting, the volunteer actors can let the system give feedback for the unofficial practice sessions, and therefore students get more practice with feedback. Also, the lecturer could study the conversations of the top performers and choose interesting segments to discuss. As student doctors get better in rapport building, when they graduate and practice as doctors, treatments are more effective and long-term \citep{egbert1964reduction, dimatteo1979social, travaline2005patient}. 

Outside of the healthcare domain, understanding and extracting the features required to detect rapport-building could help researchers build better conversational systems. Our first contribution is the identification of multimodal features that have been found to be associated with rapport building and using them to predict rapport building automatically.  Our second contribution is to include them into a text-based multimodal narrative system \cite{kim2019detecting}. Why go through text? It is because this is how human experts have been manually analyzing conversations in the linguistics community. Our text-based approach has the merit of  emulating the way human analysts analyze conversations, and hence supporting better interpretability. We demonstrate that the additions bring statistically significant improvements. This feature-engineering system\footnote{Open-sourced at https://github.com/SpectData/MONAH} could potentially be used to accomplish a highly attention-demanding task for an analyst. With an automated text-based approach, we aim to contribute towards the research gap of automatic visualizations that support multimodal analysis \cite{kim2019review}. The created multimodal transcript itself is a conversational analysis product, which can be printed out on paper.

\begin{figure*}[!b]
    \includegraphics[width=\textwidth]{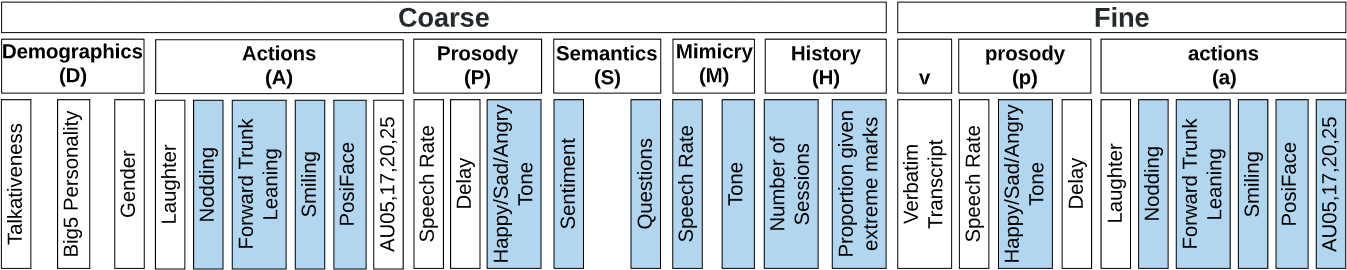}
    \caption{High-level features introduction. We build on our previous work \citep{kim2019detecting} -- the new features introduced in this work are coloured in blue, whilst the existing set of features are in white.}
    \label{fig:HighLevel}
\end{figure*}

In this paper, we first introduced the problem domain (section \ref{sect:data}). Secondly, we motivated the new features (detailed in Fig. \ref{fig:HighLevel}) to be extracted (section \ref{sec:features}). Then, we extracted the features from videos and encoded them as text together with verbatim transcripts (section \ref{sec:features}). To evaluate whether the text narratives were useful, we ran experiments that predict rapport-building using texts containing different amounts of multimodal annotations (section \ref{sect:expSetting}). Finally, we discuss the results and visualize the outputs of the system (section \ref{sect:expresults}).

\section{Related Works}
\label{sect:related}

The automated analysis of conversations has been the subject of considerable interest in recent years. Within the domain of doctor-patient communication, \citet{sen-etal-2017-modeling} calculated session-level input features, including affective features \cite{gilbert-2014-vader}. Analyses using session-level features have a drawback of not being able to identify specific defining multimodal interactions in the conversation \cite{zhao2016socially, heylen2007searching}. Therefore, we build upon the works of \citet{sen-etal-2017-modeling} – in addition to the use of session-level features, we propose using a finer level of talk-turn multimodal text representation as inputs into a hierarchical attention network (HAN) \cite{yang2016hierarchical}. 

We also build upon our previous work \citep{kim2019detecting} by broadening the range of multimodal features considered. As for the different methods of multimodal information fusion, \citet{poria2017review} completed an extensive review of the different state-of-the-art multimodal fusion techniques. Recent multimodal fusion research (such as ICON \citep{hazarika2018icon}, CMN \citep{hazarika2018conversational}, MFN \citep{zadeh2018memory}, DialogueRNN \citep{majumder2019dialoguernn}, M3ER \citep{mittal2020m3er}) has focussed on end-to-end approaches. Unlike the typical end-to-end approach of representing and fusing multimodal features using numeric vectors, our contribution is an entirely text-based multimodal narrative, thereby improving downstream analysis's interpretability. The approach of this system not only annotates the presence of nonverbal events \cite{eyben2011string}, but also the degree of the nonverbal event intensity at both the session-level and talkturn-level.

\section{Data}
\label{sect:data}
This study uses data from the EQClinic platform \cite{liu-etal-2016-webbased}. Students in an Australian medical school were required to complete at least one medical consultation on the online video conferencing platform EQClinic with a simulated patient who is a human actor trained to act as a patient. Each simulated patient was provided with a patient scenario, which mentioned the main symptoms experienced. The study was approved by the Human Research Ethics Committee of the University of New South Wales (project number HC16048).

The primary outcome measurement was the response to the rapport-building question on the Student-Patient Observed Communication Assessment (SOCA) form, an adapted version of the Calgary-Cambridge Guide \cite{kurtz1996calgary}. Simulated patients used the SOCA form to rate the students’ performances after each video consultation. Our dataset comprises of 873 sessions, all from distinct students. Since we have two recordings per session (one of the student, the second of the simulated patient), the number of recordings analyzed is 1,746. The average length per recording is 928 seconds (sd=253 seconds), amounting to a total of about 450 hours of recordings analyzed. The dataset's size is small relative to the number of multimodal features extracted; therefore, there is a risk of overfitting.

We used the YouTube platform to obtain the transcript per speaker from the recordings. We chose YouTube because we  \citep{kim2019comparison} found that it was the most accurate transcription service (word error rate: 0.28) compared to Google Cloud (0.34), Microsoft Azure (0.40), Trint (0.44), IBM Watson (0.50), when given dyadic video-conferences of an Australian medical school. \citet{jeong2020analysis} found that among the four categories of YouTube errors (omission, addition, substitution, and word order), substitution recorded the highest amount of errors. Specifically, they found that phrase repetitions could be mis-transcribed into non-repetitions. From our experience, (a) repair-initiation techniques such as sound stretches (e.g. ``ummmm") \citep{hosoda2006repair}, were either omitted or substituted with ``um"; (b) overlapping speech was not a problem because our speakers were physically separated and recorded into separate files.

We brought together the two speakers’ transcripts into a session-level transcript through word-level timings and grouped together words spoken by one speaker until the sequence is interrupted by the other speaker. When the interruption occurs, we deem that the talk-turn of the current speaker has ended, and a new talk-turn by the interrupting speaker has begun. The average number of talk-turns per session is 296 (sd=126), and the average word count per talk-turn is 7.62 (sd=12.2). 

At this point, we note that acted dialogues differ from naturally occurring dialogues in a few ways. Firstly, naturally occurring dialogues tend to be more vague (phrases like ``sort of", ``kinda", ``or something") due to the shared understanding between the speakers \citep{quaglio2008television}. Secondly, taboo words or expletives that convey emotions (like ``shit", ``pisssed off", ``crap") is likely to be less common in an acted medical setting than naturally occurring conversations. Some conversations transform into genuine dialogues where the speakers ``shared parts of themselves they did not reveal to everyone and, most importantly, this disclosure was met with acceptance" \citep{montague2012genuine}. This definition of genuine conversation is similarly aligned to our definition of rapport-building in section \ref{ssec:depvar}.

\begin{table*}[!b]
  \centering
  \caption{Session-level input features for each participant. \text{*} indicates new features outside of \citet{kim2019detecting}.}
  \input{Tables/featuredescription}
  \label{tab:featuredescription}
\end{table*}

\begin{figure}[h!]
    \centering
    \includegraphics[width=0.50\textwidth]{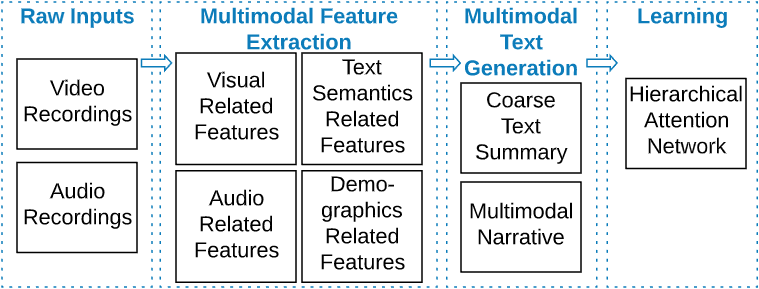}
    \caption{MONAH (Multi-\underline{Mo}dal \underline{Na}rratives for \underline{H}umans) Framework.}
    \label{fig:trainingframework}
\end{figure}

\autoref{fig:HighLevel} shows a summary of the features extracted. We annotated verbatim transcripts with two different levels of multimodal inputs – annotations at the session-level are labeled \textit{coarse}, whilst annotations at the talk-turn-level are labeled \textit{fine}. To facilitate comparisons, all input families belonging to the \textit{coarse} (\textit{fine}) level would be annotated with uppercase (lowercase) letters, respectively. In this paper, we refer to the previously existing set of features (with white background) as the “prime” (\textprime) configuration. Families are also abbreviated by their first letter. For example, the \textit{coarse} $P\textprime$ family would consist of only speech rate and delay, whilst the \textit{coarse} $P$ family would consist of $P\textprime$ plus tone. As another example, the \textit{coarse} $D\textprime$ family is the same as the $D$ family because there are no newly added features (in blue). We introduce the framework of our multimodal feature extraction pipeline in \autoref{fig:trainingframework}.

\section{Multimodal features extractions}
\label{sec:features}

As an overview, we extracted the timestamped verbatim transcripts and used a range of pre-trained models to extract temporal, modality-specific features. We relied on pre-trained models for feature extraction and did not attempt to improve on them -- demonstrating the value of using multidisciplinary pre-trained models from natural language processing, computer vision, and speech processing for conversational analysis.

Effectively, we extracted structured data from unstructured video data (section \ref{ssec:descFeatures}). With the structured data and verbatim transcript, we weaved a multimodal narrative using a set of predefined templates (sections \ref{ssec:genCoarse} and \ref{ssec:genFine}). With the multimodal narrative, we employed deep learning techniques and pre-trained word embeddings to predict the dependent variable (section \ref{sect:expSetting}).

\subsection{Dependent variable - rapport building}
\label{ssec:depvar}
The dependent variable is defined as the success in rapport building. Rapport building is one of the four items scored in the SOCA. The original 4-point Likert scale is {Fail, Pass-, Pass, Pass+}, we converted this scale into a binary variable where it is true if the rapport-building score is “Pass+” as we are concerned here with identifying good rapport building. “Pass+” means that the actor felt rapport such that all information could be comfortably shared. 38 percent of the population has achieved “Pass+”. All actors followed the same pre-interview brief. Because only the actor scored the student performance and there is no overlap, the limitation is that we do not have measures of agreement.

\subsection{Description of features}
\label{ssec:descFeatures}

Table \ref{tab:featuredescription} gives an overview of all features for each speaker. We define six families of \textit{coarse}-level inputs -– \textit{demographics}, \textit{actions}, \textit{prosody}, \textit{semantics}, \textit{mimicry}, and \textit{history}. We computed the features per speaker. From all families, there are a total of 77 features per session. 

We first discuss the family of \textit{demographics}. \textit{Talkativeness} is chosen because the patient’s talkativeness would initiate the doctor’s active listening while aiding identification of patient’s concerns – processes that could establish rapport. In \citet{hall2009observer}, it appears that patients appreciate a certain degree of doctor’s dominance in the conversation, which itself is also correlated with higher rapport. \textit{Big 5 Personality} consists of Extraversion, Agreeableness, Conscientiousness, Neuroticism, and Openness to Experience \citep{mccrae1987validation}. This personality structure is widely used in research and practice to quantify aspects of a person's natural tendency in thought, feeling, and action, with good validity and reliability indicators \citep{mccrae2017five}. It is chosen because traits of agreeableness and openness on the part of both doctor and patient predict higher rapport. Among doctors, higher openness and agreeableness predict higher empathy towards patients \cite{costa2014associations}. Among patients, higher agreeableness predicted higher trust towards doctors \cite{cousin2013agreeable}, and higher openness predicted higher doctor affectionate communication \cite{hesse2019relationships}. \textit{Big 5 Personality} is extracted through feeding transcripts to the IBM Watson Personality Insights API (version 2017-10-13), costing a maximum of 0.02 USD per call. \textit{Gender} is chosen because personality differences between genders were observed cross-culturally. Among twenty-three thousand participants across cultures for both college-age and adult samples, females reported higher agreeableness, warmth, and openness to feelings than males \cite{costa2001gender}, traits that could be linked to rapport building.

Secondly, for the family of \textit{actions}, \textit{laughter} is chosen because humor (which was defined in part by the presence of laughter) on the part of both doctor and patient was found to be twice as frequent in high-satisfaction than low-satisfaction visits \cite{sala2002satisfaction}. Laughter events were detected using the \citet{ryokai2018capturing} algorithm. \textit{Facial expressions} that resemble smiling is another behavioral indicator of humor appreciation, and approval of one another \cite{tickle1990nature}. \textit{Head nodding} is a type of backchannel response (i.e., response tokens) that has been shown to reflect rapport between doctor and patient, especially when the primary activity is face to face communications \cite{manusov2014sourcebook}. \textit{Forward trunk leaning} is chosen because it has long been found to reflect an expression of interest and caring, which are foundational to rapport building \cite{scheflen1964significance}. Additionally, facial positivity (\textit{posiface}) is included as it is useful in rapport building detection in small groups \cite{muller2018detecting}. Lastly, \textit{action units} (AU) that describe specific facial expressions, in particular AU 05 (upper lid raiser), 17 (chin raiser), 20 (lip stretcher), 25 (lips part), are also included as they were useful in automated dyadic conversational analyses to detect depression in our previous work \cite{kim2019detecting}. All features introduced in this paragraph were calculated using the AU and landmark positioning features extracted using OpenFace \cite{baltruvsaitis2016openface}.

Thirdly, for the family of \textit{prosody}, \textit{delay} is chosen because it has been shown to be an indicator of doctor-to-patient influence -- patients of low rapport with their doctors were found to speak less in response to doctor’s comments \cite{sexton1996interaction}. \textit{Speech rate} is chosen because doctor’s fluent speech rate and patient’s confident communication have been positively correlated with the patient’s perception of rapport \cite{hall2009observer}. Delay and speech rate are calculated using the time-stamped transcripts. \textit{Tone} is chosen because a warm and respectful tone on the part of both doctor and patient is positively correlated with the patient’s perception of rapport \cite{hall2009observer}. Tone is calculated using the Vokaturi algorithm (version 3.3) \cite{vokaturi2019}.

\begin{table*}[h]
  \centering
  \caption{Templates for the session-level \textit{coarse} summary.}
  \input{Tables/coarsetemplates}
  \label{tab:coarsetemplates}
\end{table*}

Fourthly, for the family of \textit{semantics}, \textit{sentiment} is chosen because the provision of positive regard from a practitioner to a patient is an important factor to foster therapeutic alliance; additionally, this process may be further enhanced if the patient also demonstrates positive behaviors towards the practitioners \cite{farber2011positive}. Sentiment is extracted using the VADER algorithm \cite{gilbert-2014-vader}, in line with \citet{sen-etal-2017-modeling}. \textit{Questions} is chosen because higher engagement by the doctor (e.g., asking questions) with the patient and the patient asking fewer questions have been shown to positively correlate with the patient’s perception of rapport \cite{hall2009observer}. Questions are detected using Stanford CoreNLP Parser \cite{manning-EtAl:2014:P14-5} and the Penn Treebank \cite{bies1995bracketing} tag sets.

Next, \textit{mimicry} is chosen because doctor-patient synchrony is an established proxy for rapport. In a review paper, rapport is theorized to be grounded in the coupling of practitioner’s and patient’s brains \cite{koole2016synchrony}. Such a coupling process would eventuate in various forms of mimicry in the dyad, for instance, vocally (e.g., matching speech rate and tone), physiologically (e.g., turn-taking, breathing), physically (e.g., matching body language) \cite{wu2020automatic}. In this study, we aim to use vocal mimicry to capture this underlying phenomenon. Session level mimicry scores are approximated through Dynamic Time Wrapping distances \cite{giorgino2009computing}, in line with \citet{muller2018detecting}.

Lastly, \textit{history} is chosen because the scores given by the assessors could be subjective evaluations where the evaluations are unduly influenced by the assessor’s leniency bias \cite{moers2005discretion}. We attempted to mitigate the leniency bias by introducing history features that indicate the assessor’s leniency and its consistency.

\subsection{Generation of \textit{coarse} multimodal narrative}
\label{ssec:genCoarse}
In this section, we discuss the \textit{coarse} multimodal narrative. We summarized the automatic generation of the text representation in Table \ref{tab:coarsetemplates}.

\begin{table*}[ht]
  \centering
  \caption{Templates for the talkturn-level \textit{fine} summary.}
  \input{Tables/finetemplates}
  \label{tab:finetemplates}
\end{table*}

We calculated the z-score for all the above templates (except Template 3 which is categorical) using the following z-score formula. The average ($\mu$), and standard deviation ($\sigma$) are computed using observations from the training observations. Using the z-score, we bucketed them into ``very low" (z$<$-2), ``low" (z$<$-1), ``high" (z$>$1) and ``very high" (z$>$2). The reason for z-transformation is to create a human-readable text through bucketing continuous variables into easy-to-understand buckets (``high" vs. ``low").

\begin{equation}
    \centering
    z = \frac{x - \mu_{\text{Train}}}{\sigma_{\text{Train}}}
    \label{eq:1}
\end{equation}

\subsection{Generation of \textit{fine} multimodal narrative}
\label{ssec:genFine}
In addition to the verbatim transcript, we introduced two new families of information – \textit{prosody}, and \textit{actions}. Table \ref{tab:finetemplates} gives an overview of the templates, and the bold-face indicates a variable. The motivations of the features have been discussed; we discuss the rules of insertion in the next few paragraphs.

Template 19 is the verbatim transcript returned from the ASR system. Before each talk-turn, we identified the speaker (doctor/patient) and added multimodal information using templates 20-29. Speech rate and tone were standardized across all training observations. We appended template 20, 21 where possible values are dependent on the z-score – “quickly” (1 $<$ z-score $<$ 2) and “very quickly” (z-score $\geq$ 2). For delay, we used time intervals of 100 milliseconds, and between 200 and 1200 milliseconds -- in line with \citet{roberts2013identifying}. We appended template 22 at the front of the talk-turn if a delay of at least 200 milliseconds is present between talk-turns.  In addition, we appended template 23 where possible values are dependent on the standardized duration of delay – “short” ($<$ 1 z-score), “long” ($<$ 2 z-score) and “significantly long” ($\geq$ 2 z-score). Template 23 captures longer than usual delay, considering the unique turn-taking dynamics of each conversation. The standardized duration of delay is calculated using talk-turn delays from the respective session. Lastly, as for the actions family, templates 24 – 28 were added if any of the actions are detected during the talk-turn. For template 29, it was only added if the AU is detected throughout the entire duration of the talk-turn.

\section{Experimental settings}
\label{sect:expSetting}

There are two main types of inputs – (1) numeric inputs at the session-level, and (2) \textit{coarse} and/or \textit{fine} multimodal narrative text inputs. As an overview, for (1), we trained the decision tree classifier using session-level numeric inputs. As for (2), we trained the HAN \cite{yang2016hierarchical}. We aim to facilitate how humans analyze conversations -- HAN can work with text and has easy interpretation with single-headed attention, making it a suitable candidate. Relative to BERT \citep{devlin2018bert}, the HAN is faster to train and easier to interpret.

\subsection{Research questions}
\label{ssec:researchquestions}

\begin{table*}[ht]
  \centering
  \caption{Summary of the model performances. We report the average five-fold cross-validation AUC and its standard deviation in brackets. Row-wise: We begin with the $D\textprime A\textprime P\textprime$, which is the full existing feature set from \citet{kim2019detecting}, and progressively compare it against the new sets of features to answer \textbf{Q1}. Column-wise: We compare the difference in AUC between the classification tree and \textit{coarse}-only HAN to answer \textbf{Q2}. We compare the difference in AUC between the \textit{coarse}-only HAN and \textit{coarse} + \textit{fine} HAN to answer \textbf{Q3}. Asterisks (*) indicate significance relative to the $D\textprime A\textprime P\textprime$ row. Carets (^) indicate significance relative to column-wise comparisons, we also provide the confidence intervals in square brackets [] for the difference in performance. The number of symbols  indicate the level of statistical significance, e.g., ***: 0.01, **: 0.05, *: 0.10.}
  \input{Tables/results}
  \label{tab:summarymodels}
\end{table*}

The proposed features have been motivated by scientific studies in Section \ref{sec:features}. A natural next question is, ``what are the impacts of these proposed features on model performance?" We break this broad question into three questions.

Firstly, (\textbf{Q1}) do the newly added features improve performance over the existing set of features for the classification tree and/or HAN?

Secondly, modelling using unstructured text input data (as opposed to using numeric inputs) has the risk of introducing too much variability in the inputs. Therefore,  we investigate (\textbf{Q2}) -- given the \textit{coarse}-only inputs, do the performance between the HAN and classification tree differ significantly?

Lastly, adding more granular talkturn-level inputs to the \textit{coarse} session-level inputs has the benefit of deeper analyses, because it allows the analyst to analyze important talkturns of the conversation. On top of this benefit, (\textbf{Q3}) do we also have significant performance improvement between \textit{coarse}-only vs. both \textit{coarse} and \textit{fine} inputs?

For all models, the area under the receiver-operator curve (AUC) was used as the evaluation metric. The AUC measures the goodness of ranking \citep{hanley1982meaning} and therefore does not require an arbitrary threshold to turn the probabilities into classes. The partitioning of the dataset to the five-folds is constant for decision tree and HAN to facilitate comparison. The five folds are created through stratified sampling of the dependent variable.

\subsection{Classification tree set-up}
\label{ssec:analysistree}

To answer (\textbf{Q1}) and (\textbf{Q2}), we tested for all 72 configurations of prime ($2^3=8$) plus full ($2^6=64$) family inputs for the decision tree.  We performed the same z-transformation pre-processing (as in section \ref{ssec:genCoarse}) on the decision tree input variables and limited random search to twenty trials.

The algorithm used is from the \textit{rpart} package with R. As part of hyperparameter tuning, we tuned the cp (log-uniform between $10^{-7}$ to $10^{-9}$), maximum depth (uniform between 1 to 20), and minimum split (uniform between 20 to 80) through five-fold cross-validation and random search.

\subsection{HAN set-up}
\label{ssec:analysishan}

To answer (\textbf{Q1}) and (\textbf{Q2}), we chose the input configurations that performed that best for the classification tree, and used the same input configurations in HAN to compare the difference. Therefore, this test is biased in favour of the classification tree. To answer (\textbf{Q3}), we added the \textit{fine} narratives to each \textit{coarse}-only configuration, and compared the difference.

The model architecture is the HAN architecture by \citet{yang2016hierarchical}, with about 5 million parameters. We used the pre-trained Glove word embeddings \cite{pennington2014glove} of 300-dimensions to represent each word. Words not found in the Glove vocabulary are replaced with the “unk” token. The hyperparameter tuning procedure is reported in Appendix \ref{ssect:hyperparam_range}, and the best hyperparameter configurations are reported in Appendix \ref{ssect:AppendixA}. There are twenty hyperparameter search trials for each input configuration\footnote{We conducted additional tuning experiments for the tree in Appendix \ref{ssect:AppendixB} to observe potential improvements in performance.}.

\section{Experimental results}
\label{sect:expresults}

The results are summarized in Table  \ref{tab:summarymodels}. The key findings are: (\textbf{Q1}) with the extended inputs, we observed statistically significant improvements in both the HAN and tree over the existing full set of features (one-tailed \textit{t}-test); (\textbf{Q2}) given the \textit{coarse}-only inputs, the performances between the HAN and classification tree did not differ significantly (two-tailed \textit{t}-test), therefore it is plausible that feature engineering into text features do not risk performance; (\textbf{Q3}) although adding the \textit{fine} narratives allow deeper analyses by the analyst, it does not lead to significant differences over the \textit{coarse}-only inputs (two-tailed \textit{t}-test). 

(\textbf{Q1}) When compared to the full set of existing features, the classification tree achieved statistically significant improvements (at $\alpha = 0.05$) in all six out of six \textit{coarse} input families. For HAN, it achieved statistically significant improvements in one (at $\alpha = 0.05$) or two (at $\alpha = 0.10$) out of six \textit{coarse} input families. This demonstrates the value of the newly introduced \textit{coarse} features\footnote{We performed additional tests in Appendix \ref{ssect:additional_fine} to observe the impact of the additions to the \textit{fine} narratives, and found small improvements (but statistically insignificant) in all three out of three input families ($va$, $vp$, $vpa$).}.

(\textbf{Q2}) Across the seven \textit{coarse} input configurations, there are no significant differences in the performance from the classification tree when compared to the HAN in six out of seven input configurations. The only exception is in the baseline $D\textprime A\textprime P\textprime$ configuration where the HAN is significantly better. However, the lack of statistically significant differences does not mean that the performances are the \textit{same}. In line with \citet{quertemont2011statistically} recommendation, we provided the confidence interval around the difference in performance for discussion. Of all confidence intervals that included zero in the fourth column of Table \ref{tab:summarymodels}, the confidence intervals do not suggest that that the effect sizes are negligible (for example, less than 0.01). In summary, we cannot conclude that the performance of HAN differs significantly from tree nor are they the same.

(\textbf{Q3}) The addition of \textit{fine} narratives to the \textit{coarse} narrative did not result in significantly stronger (nor weaker) performance in any of the seven input configurations. We posit that this negative finding is due to the difficulty in prioritizing the back-propagation updates to the parts of the network interacting with the \textit{coarse} features, where there is likely a high signal-to-noise ratio. Despite the negative finding, we think it is important to explore fine features' addition onto coarse features because it produces a complete transcript for the human to understand how the conversation proceeded.

\subsection{Qualitative Analysis}
\label{ssect:qualitativeanalysis}
We visualized the talkturn-level and word-level attention weights from the model. Attention weights are normalized using z-transformation and bucketed into four buckets ($<0$, $<1$, $<2$, $\geq$ 2) \cite{kim2019detecting}. The analyst could analyze an important segment in detail (as in Fig. \ref{fig:conversationanalysisexample}) or see an overview of the important segments in the conversation (see appendix \ref{ssect:AppendixC}). In the example (Fig. \ref{fig:conversationanalysisexample}), we observed that the multimodal annotations of leaning forward and positive expression were picked up as important words by the model.

\begin{figure}[ht]
    \includegraphics[width=0.5\textwidth]{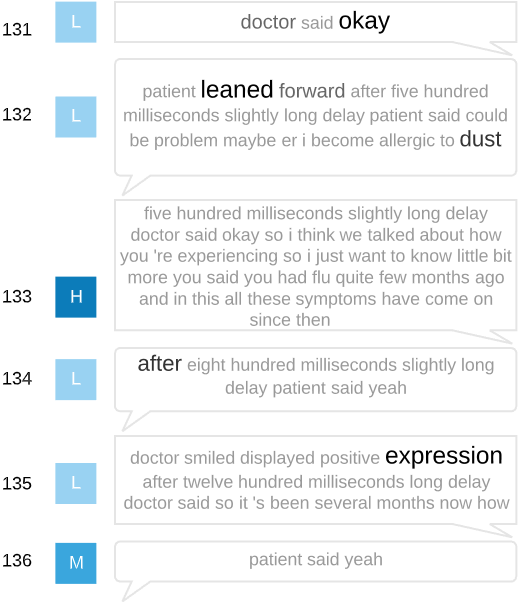}
    \caption{Conversation analysis for a true positive. The talkturn-level attentions are labelled Low (L), Medium (M) and High (H), while the words with higher attention have a larger and darker font. We also transcribed this segment using the Jefferson system in Appendix \ref{ssect:AppendixD}.}
    \label{fig:conversationanalysisexample}
\end{figure}

\section{Conclusion}
\label{sect:conclusion}

In this paper, we build upon a fully text-based feature-engineering system. We motivated the added features with existing literature, and demonstrated the value of the added features through experiments on the EQClinic dataset. This approach emulates how humans have been analyzing conversations with the \citet{jefferson2004glossary} transcription system, and hence is human-interpretable. It is highly modular, thereby allowing practitioners to inject modalities. In this paper, we have used a wide range of modalities, including \textit{demographics}, \textit{actions}, \textit{prosody}, \textit{mimicry}, \textit{actions}, and \textit{history}. The ablation tests showed that the added \textit{coarse} features significantly improve the performance for both decision tree and HAN models. 

Future research could (1) investigate whether this feature engineering system is generalizable to wider applications of conversational analysis; (2) conduct user studies to validate the usability and ease of interpretability of the visualization.

\section*{Acknowledgments}

We acknowledge the Sydney Informatics Hub and the University of Sydney’s high-performance computing cluster, Artemis, for providing the computing resources and Marriane Makahiya for supporting the data manipulation work. Video data collection was carried out as part of the OSPIA platform project, funded by the Department of Health Clinical Training Fund from the Australian Government.

\bibliography{anthology,eacl2021}
\bibliographystyle{acl_natbib}

\appendix

\section*{Appendices}
\renewcommand{\thesubsection}{\Alph{subsection}}

\begin{table*}[ht]
  \centering
  \caption{Best HAN configurations for the development set.}
  \input{Tables/hanbesthyperparam}
  \label{tab:hanconfigs}
\end{table*}

\begin{table}[ht]
  \centering
  \caption{Best Tree configurations for the development set.}
  \input{Tables/treebesthyperparam}
  \label{tab:treeconfigs}
\end{table}

\subsection{Tuning procedure}
\label{ssect:hyperparam_range}

We tuned the SGD optimizer with a learning rate between 0.003 to 0.010, batch size to be between 4 to 20, L2 regularization between $10^{-6}$ and $10^{-3}$, and trained for up to 350 epochs without early stopping. We tuned the number of gated recurrent units (GRU) \cite{cho2014learning} between 40 to 49 in both the word-level and talk-turn-level layers, with both the GRU dropout and recurrent dropout \cite{gal2016theoretically} to be between 0.05 to 0.50. The method of choosing hyperparameters is through uniform sampling between the above-mentioned bounds, except for the learning rate where log-uniform sampling is used. Training is performed on a RTX2070 GPU or V100 GPU.

\subsection{Hyperparameter configurations for best-performing models}
\label{ssect:AppendixA}
Table \ref{tab:hanconfigs} (HAN) and Table \ref{tab:treeconfigs} (Tree) report the hyperparameter configurations for each of the best-performing model reported in Table \ref{tab:summarymodels}.

\subsection{Performance of additional tuning}
\label{ssect:AppendixB}

\begin{figure}[ht]
    \includegraphics[width=0.5\textwidth]{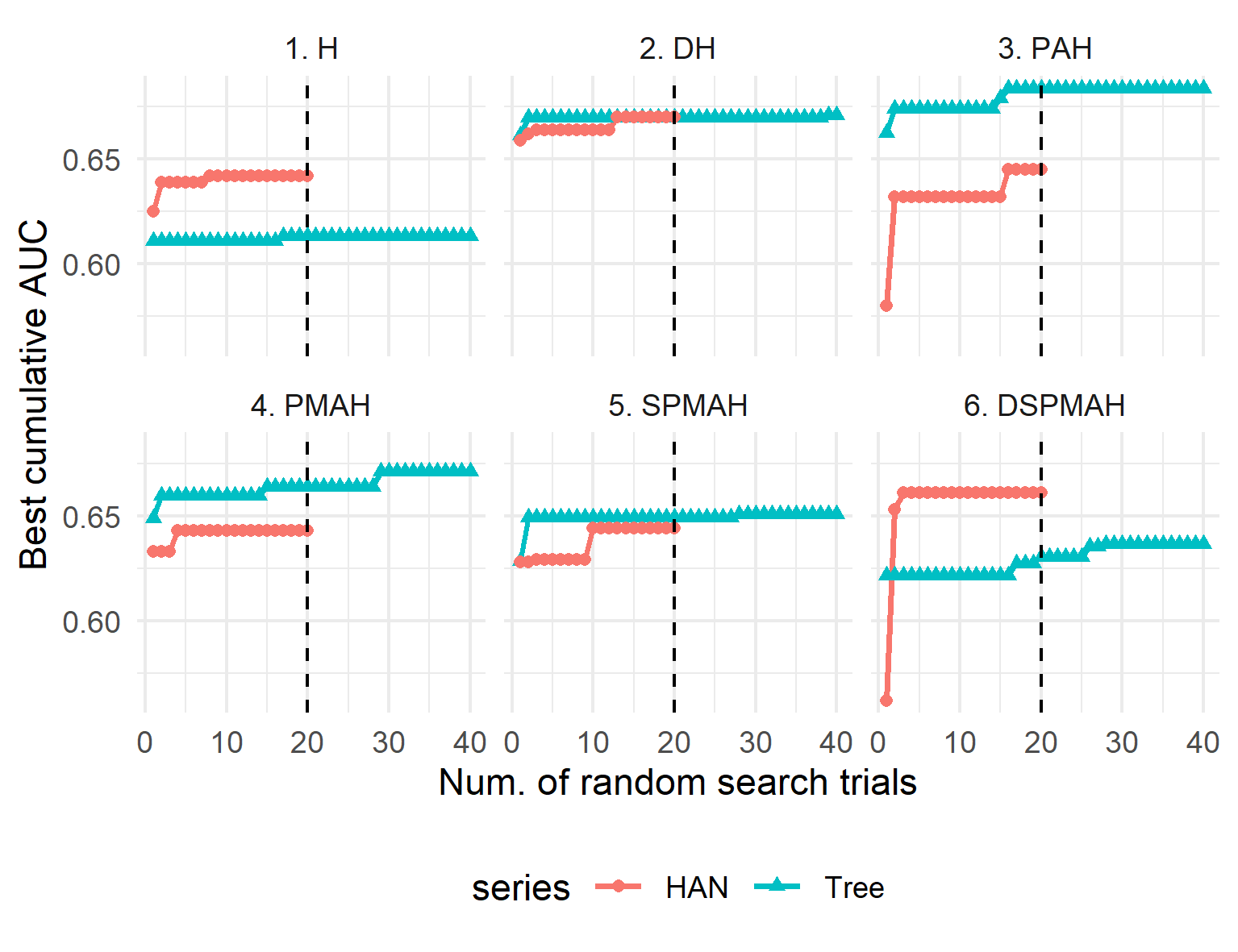}
    \caption{Best cumulative AUC performance given N random search trials.}
    \label{fig:timeperformance}
\end{figure}

We conducted additional experiments on the tree configurations to (1) compare the improvements in performance when tuning the HAN and tree, and (2) evaluated the increase in performance if the tree is allowed twenty more hyperparameters random search trials (Fig. \ref{fig:timeperformance}).

From the larger increases in HAN performances, it is plausible that HAN is more sensitive to the hyperparameter tuning than the tree.

\subsection{Additional tests for additions to the \textit{fine} narratives}
\label{ssect:additional_fine}

Table \ref{tab:additional_fine} reports the additional tests on the impact of the added \textit{fine} features. We observe that whilst all three input configurations ($va$, $vp$, $vpa$) have small increases in performance, none of them are statistically significant.

\begin{table}[ht]
  \centering
  \caption{Summary of the model performances for the fine narratives. We report the average five-fold cross-validation AUC and its standard deviation in brackets. Row-wise, we begin with the $v$ configuration to show the impact of \textit{fine} multi-modal annotations over the verbatim transcript. Then, we show the impact of the additions (\textbf{Q1}) over the existing fine annotations from \citet{kim2019detecting} using column-wise comparisons. Asterisks (*) indicate significance relative to the $v$ row. Carets (^) indicate significance relative to column-wise comparisons, we also provide the confidence intervals in square brackets [] for the difference in performance. The number of symbols  indicate the level of statistical significance, e.g., ***: 0.01, **: 0.05, *: 0.10.}
  \input{Tables/results_fine}
  \label{tab:additional_fine}
\end{table}

\subsection{Conversation thumbnail visualization}
\label{ssect:AppendixC}

\begin{figure}[ht]
    \includegraphics[width=0.5\textwidth]{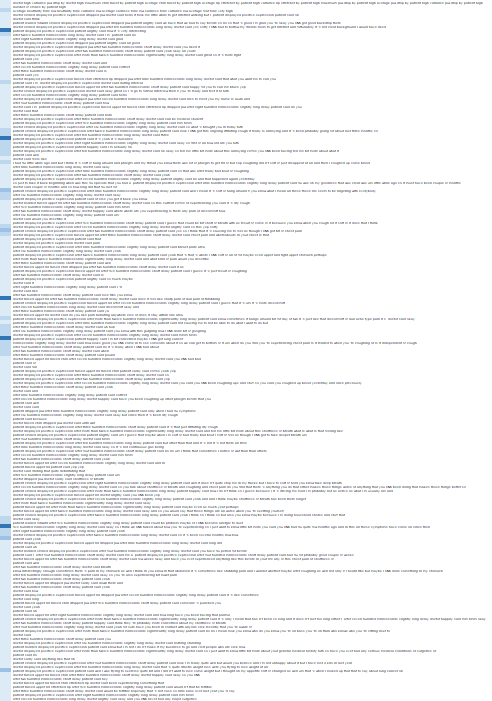}
    \caption{Heatmap thumbnail. Darker blue indicates higher talkturn attention weights.}
    \label{fig:heatmapoverview}
\end{figure}

By illustrating the talkturn-level attention weights as a heatmap thumbnail (Fig. \ref{fig:heatmapoverview}), the analyst could quickly get a sense of the important segments of the conversation without reading the content and zoom-in if required.

\subsection{Jefferson example}

\label{ssect:AppendixD}
\begin{figure}[h!]
    \includegraphics[width=0.5\textwidth]{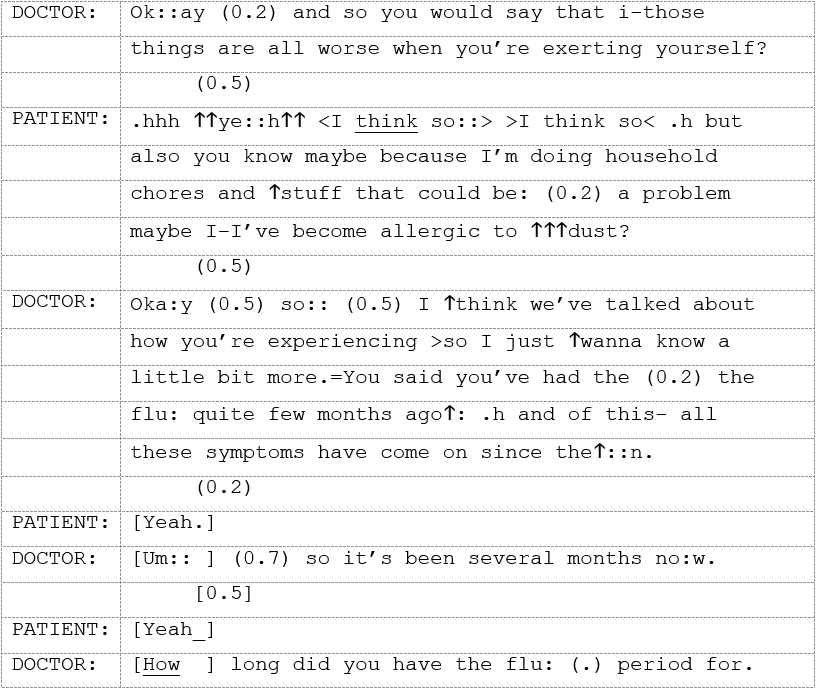}
    \caption{Jefferson transcription example. : (colon) - stretched sound; (0.2) - a pause of 0.2 seconds; .hhh - in breath, .h - short in breath; $\uparrow$ - Rise in intonation; underline - emphasis; $<>$ - slowed speech rate, $><$ - quickened speech rate; $[]$ - overlapping speech.}
    \label{fig:jefferson}
\end{figure}

As an optional reference, we engaged a professional transcriptionist to transcribe the conversation segment presented (Fig. \ref{fig:conversationanalysisexample}) using the Jefferson system. The Jefferson example is presented in Fig. \ref{fig:jefferson}. The verbal content is slightly different due to (1) different methods to determine talkturns transitions and (2) automatic speech recognition accuracy.

\newpage

\end{document}

%% file: Tables/featuredescription.tex
  \begin{tabular}{|>{\arraybackslash}m{1.40cm}
                  |>{\arraybackslash}m{2.75cm}
                  |>{\arraybackslash}m{10.75cm}|}
\hline
\textbf{Family}& \textbf{Child}& \textbf{Template}\\
\hline
\multirow{3}*{\shortstack[l]{Demo\\ graphics}}
    &Talkativeness 
    &Total word count, total distinct word count, and proportion of word count \\ \cline{2-3}
    
    &Big 5 Personality
    &Percentile scores for each of the big 5 personality \\ \cline{2-3}

    &Gender
    &Male or Female \\ \hline
\multirow{6}{*}{Actions} 
    &Laughter
    &Total laughter count \\ \cline{2-3}
    
    &Head Nodding\text{*}
    &Count of nods \\ \cline{2-3}

    &Forward Trunk Leaning\text{*}
    &Count of leaning in \\ \cline{2-3}
    
    &Smiling\text{*}
    &Count of smiles \\ \cline{2-3}
    
    &PosiFace\text{*}
    &Counts of times of positive and negative facial expressions \\ \cline{2-3}

    &AU
    &Summary statistics of the selected AU (05,17,20,25) intensities \\ \cline{2-3} \hline
\multirow{3}*{Prosody}
    &Delay
    &Summary statistics of time gaps between talk-turns \\ \cline{2-3}
    
    &Speech rate
    &Average speech rate \\ \cline{2-3}

    &Tone\text{*}
    &Happy, sad, angry tone \\ \hline
    
\multirow{2}*{Semantics}
    &Sentiment\text{*}
    &Composite, positive, neutral, and negative sentiment \\ \cline{2-3}
    
    &Questions\text{*}
    &Proportion of talk-turns that are open/closed questions \\ \hline
    
\multirow{2}*{Mimicry}
    &Speech Rate\text{*}
    &Dynamic time wrapping distance for speech rate \\ \cline{2-3}
    
    &Tone\text{*}
    &Dynamic time wrapping distance for tone \\ \hline
\multirow{2}*{History}
    &Num. Sessions\text{*}
    &Number of past sessions the assessor has scored before this \\ \cline{2-3}
    
    &Proportion given extreme marks\text{*}
    &Proportion of past sessions that the assessor has given an extreme score \\ \hline

\end{tabular}

%% file: Tables/coarsetemplates.tex
\begin{tabular}{|>{\arraybackslash}m{1.40cm}
                  |>{\arraybackslash}m{2.75cm}
                  |>{\arraybackslash}m{0.25cm}
                  |>{\arraybackslash}m{9.5cm}|}
\hline
\textbf{Family}& \textbf{Child}& \textbf{ID}& \textbf{Template}\\
\hline
\multirow{3}*{\shortstack[l]{Demo\\ graphics}}
    &Talkativeness 
    &1
    &\textbf{doctor} number of words \textbf{high}, \textbf{doctor} number of distinct words high \\ \cline{2-4}
    
    &Big 5 Personality
    &2
    &\textbf{doctor} openness \textbf{high} \\ \cline{2-4}

    &Gender
    &3
    &The \textbf{patient} is \textbf{female} \\ \hline
\multirow{6}{*}{Actions} 
    &Laughter
    &4
    &\textbf{doctor} laughter counts \textbf{high} \\ \cline{2-4}
    
    &Head Nodding
    &5
    &\textbf{doctor} head nod counts \textbf{high} \\ \cline{2-4}

    &Forward Trunk Leaning
    &6
    &\textbf{doctor} forward trunk leaning \textbf{high} \\ \cline{2-4}
    
    &Smiling
    &7
    &\textbf{doctor} smiling counts \textbf{high} \\ \cline{2-4}
    
    &PosiFace
    &8
    &\textbf{doctor} positive face expression counts \textbf{high} \\ \cline{2-4}

    &AU
    &9
    &\textbf{doctor} minimum lip depressor \textbf{very low}, maximum lip depressor \textbf{low}, average lip depressor \textbf{low}, variance lip depressor \textbf{low} \\ \cline{2-4} \hline
\multirow{3}*{Prosody}
    &Delay
    &10
    &minimum delay \textbf{very low}, maximum delay \textbf{low}, average delay \textbf{low}, variance delay \textbf{low} \\ \cline{2-4}
    
    &Speech rate
    &11
    &speech rate \textbf{high} \\ \cline{2-4}

    &Tone
    &12
    &angry tone \textbf{high} \\ \hline
    
\multirow{2}*{Semantics}
    &Sentiment
    &13
    &positive sentiment \textbf{high} \\ \cline{2-4}
    
    &Questions
    &14
    &open questions \textbf{high} \\ \hline
    
\multirow{2}*{Mimicry}
    &Speech Rate
    &15
    &speech rate mimicry \textbf{high} \\ \cline{2-4}
    
    &Tone
    &16
    &tone mimicry \textbf{high} \\ \hline
\multirow{2}*{History}
    &Num. Sessions
    &17
    &patient number of sessions before this \textbf{very high} \\ \cline{2-4}
    
    &Proportion given extreme marks
    &18
    &patient question four proportion given maximum marks \textbf{high} \\ \hline

\end{tabular}

%% file: Tables/finetemplates.tex
\begin{tabular}{|>{\arraybackslash}m{1.40cm}
                |>{\arraybackslash}m{3.50cm}
                |>{\arraybackslash}m{0.25cm}
                |>{\arraybackslash}m{8.75cm}|}
\hline
\textbf{Family}& \textbf{Child}& \textbf{ID}& \textbf{Template}\\
\hline
Verbatim
    &Transcript
    &19
    &Transcript returned from the ASR system \\ \hline
    
\multirow{3}*{Prosody}
    &Speech rate
    &20
    &the doctor \textbf{quickly} said  \\ \cline{2-4}
    
    &Tone
    &21
    &the doctor said \textbf{angrily} \\ \cline{2-4}

    &\multirow{2}*{Delay}
        &22
        &after \textbf{two} hundred milliseconds \\ \cline{3-4}
            &&23
            &a \textbf{long} delay \\ \hline
    
\multirow{6}*{Actions}
    &Laughter
    &24
    &the doctor \textbf{laughed}   \\ \cline{2-4}
    
    &Nodding
    &25
    &the doctor \textbf{nodded}   \\ \cline{2-4}

    &Forward trunk learning
    &26
    &the doctor \textbf{leaned forward}  \\ \cline{2-4}

    &Smiling
    &27
    &the doctor \textbf{smiled}  \\ \cline{2-4}

    &PosiFace
    &28
    &the doctor \textbf{displayed positive facial expression}  \\ \cline{2-4}

    &AU05, 17, 20, 25
    &29
    &the doctor \textbf{exhibited lip depressor}  \\ \hline

\end{tabular}

%% file: Tables/results.tex
\begin{tabular}{|>{\centering\arraybackslash}m{3.00cm}
                |>{\centering\arraybackslash}m{1.50cm}
                |>{\centering\arraybackslash}m{1.50cm}
                |>{\centering\arraybackslash}m{2.50cm}
                |>{\centering\arraybackslash}m{1.90cm}
                |>{\centering\arraybackslash}m{2.70cm}|}
\hline
            {\textbf{Coarse Inputs}}     &   {\textbf{Tree}}
        &   {\textbf{Coarse-only (HAN)}}   &   {\textbf{Significance of Difference (Coarse-only vs. Tree)}}
        &   {\textbf{Coarse + Fine (HAN)}} &   {\textbf{Significance of Difference (Coarse + Fine vs. Coarse-only)}} \\
\hline

            {$D\textprime A\textprime P\textprime$ 
            \newline (Existing Features)}     &{0.577 (0.011)} &{0.637 (0.018)} &{^^^\newline [0.038, 0.082]} &{0.629 (0.041)} &{ [-0.054, 0.038]} \\
\hline
            {$H$}     &{0.613 ** (0.036)} &{0.642 (0.038)} &{[-0.025, 0.083]} &{0.652 (0.048)} &{[-0.053, 0.073]} \\
\hline
            {$DH$}     &{0.670 *** (0.049)} &{0.670 ** (0.034)} &{[-0.062, 0.062]} &{0.654 (0.030)} &{[-0.063, 0.031]} \\
\hline
            {$PAH$}     &{0.684 *** (0.022)} & {0.645 (0.043)} &{[-0.089, 0.011]} &{0.661 (0.029)} &{[-0.038, 0.070]} \\
\hline
            {$APMH$}     &{0.664 *** (0.037)} & {0.643 (0.036)} &{[-0.074, 0.032]} &{0.657 (0.037)} &{[-0.039, 0.067]} \\
\hline
            {$APSMH$}     &{0.649 *** (0.021)} &{0.644 (0.049)} &{[-0.060,  0.050]} &{0.653 (0.051)} &{[-0.064, 0.082]} \\
\hline
            {$DAPSMH$}     &{0.630 *** (0.032)} &{0.661 * (0.030)} &{[-0.014, 0.076]} &{0.650 (0.028)} &{[-0.053, 0.031]} \\
\hline
\end{tabular}

%% file: Tables/hanbesthyperparam.tex
\begin{tabular}{|>{\centering\arraybackslash}m{2.90cm}
                |>{\centering\arraybackslash}m{0.90cm}
                |>{\centering\arraybackslash}m{0.90cm}
                |>{\centering\arraybackslash}m{1.40cm}
                |>{\centering\arraybackslash}m{1.20cm}
                |>{\centering\arraybackslash}m{1.60cm}
                |>{\centering\arraybackslash}m{2.35cm}
                |>{\centering\arraybackslash}m{0.90cm}|}
\hline
            {\textbf{Config.}}     &   {\textbf{Batch Size}}
        &   {\textbf{Num. of GRU}}  
        &   {\textbf{Learning Rate}}     &   {\textbf{GRU dropout}}
        &   {\textbf{GRU recurrent dropout}} &   {\textbf{L2 regularization}} &   {\textbf{Epoch}} \\
\hline

{$H$}      & {19} & {42}  & {0.010} & {0.10} & {0.23} &   {$1 \times 10^{-4}$} & {223} \\ \hline
{$DH$}     & {11} & {46}  & {0.010} & {0.07} & {0.09} &  {$3 \times 10^{-6}$}  & {74} \\ \hline
{$PAH$}     & {14} & {47}  & {0.005} & {0.16} & {0.50} & {$2 \times 10^{-5}$}  & {329} \\ \hline
{$APMH$}     & {8} & {44}  & {0.005} & {0.29} & {0.16} & {$1 \times 10^{-3}$}  & {275} \\ \hline
{$APSMH$}     & {9} & {43}  & {0.005} & {0.16} & {0.48} & {$4 \times 10^{-5}$} & {305} \\ \hline
{$DAPSMH$}     & {14} & {41}  & {0.010} & {0.49} & {0.48} & {$2 \times 10^{-5}$} & {138} \\ \hline
{$D\textprime A\textprime P\textprime$}      & {19} & {46}  & {0.004} & {0.06} & {0.50} & {$1 \times 10^{-4}$} & {260} \\ \hline

{$v$}      & {16} & {40}  & {0.009} & {0.15} & {0.09} &   {$2 \times 10^{-5}$} & {316} \\ \hline
{$va$}      & {13} & {43}  & {0.007} & {0.13} & {0.48} &   {$1 \times 10^{-6}$} & {347} \\ \hline
{$vp$}      & {8} & {42}  & {0.006} & {0.13} & {0.05} &   {$2 \times 10^{-5}$} & {310} \\ \hline
{$vpa$}      & {9} & {48}  & {0.010} & {0.45} & {0.46} &   {$1 \times 10^{-5}$} & {349} \\ \hline
{$v a\textprime$}      & {12} & {40}  & {0.006} & {0.11} & {0.30} &   {$1 \times 10^{-4}$} & {346} \\ \hline
{$v p\textprime$}      & {11} & {42}  & {0.007} & {0.44} & {0.19} &   {$2 \times 10^{-5}$} & {341} \\ \hline
{$v p\textprime a\textprime$}      & {10} & {45}  & {0.008} & {0.31} & {0.41} &   {$4 \times 10^{-6}$} & {267} \\ \hline

{$H$-$vpa$}      & {8} & {42}  & {0.005} & {0.38} & {0.33} &   {$2 \times 10^{-5}$} & {346} \\ \hline
{$DH$-$vpa$}      & {12} & {44}  & {0.009} & {0.25} & {0.14} &   {$1 \times 10^{-5}$} & {316} \\ \hline
{$PAH$-$vpa$}      & {11} & {47}  & {0.005} & {0.08} & {0.49} &   {$5 \times 10^{-5}$} & {349} \\ \hline
{$APMH$-$vpa$}      & {18} & {46}  & {0.008} & {0.13} & {0.50} &   {$1 \times 10^{-5}$} & {339} \\ \hline
{$APSMH$-$vpa$}      & {9} & {43}  & {0.010} & {0.13} & {0.21} &   {$2 \times 10^{-6}$} & {240} \\ \hline
{$DAPSMH$-$vpa$}      & {15} & {46}  & {0.009} & {0.15} & {0.50} &   {$2 \times 10^{-5}$} & {340} \\ \hline
{$D\textprime A\textprime P\textprime$ - $v p\textprime a\textprime$}      & {13} & {46}  & {0.008} & {0.26} & {0.16} &   {$1 \times 10^{-5}$} & {262} \\ \hline

\end{tabular}

%% file: Tables/treebesthyperparam.tex
\begin{tabular}{|>{\centering\arraybackslash}m{2.50cm}
                |>{\centering\arraybackslash}m{0.90cm}
                |>{\centering\arraybackslash}m{0.90cm}
                |>{\centering\arraybackslash}m{1.80cm}|}
\hline
            {\textbf{Config.}}     &   {\textbf{Min. split}}
        &   {\textbf{Max. depth}}  
        &   {\textbf{cp}} \\
\hline

{$H$}      & {27} & {17}  &  {$3.13 \times 10^{-6}$} \\ \hline
{$DH$}     & {72} & {18}  &  {$1.14 \times 10^{-6}$} \\ \hline
{$PAH$}    & {70} & {15}  &  {$8.84 \times 10^{-5}$} \\ \hline
{$APMH$}   & {72} & {18}  &  {$1.14 \times 10^{-6}$} \\ \hline
{$APSMH$}  & {68} & {14}  &  {$5.26 \times 10^{-5}$} \\ \hline
{$DAPSMH$} & {37} & {10}  &  {$2.94 \times 10^{-5}$} \\ \hline
{$D\textprime A\textprime P\textprime$}      & {68} & {21}  &  {$3.74 \times 10^{-5}$} \\ \hline

\end{tabular}

%% file: Tables/results_fine.tex
\begin{tabular}{|>{\centering\arraybackslash}m{1.20cm}
                |>{\centering\arraybackslash}m{1.25cm}
                |>{\centering\arraybackslash}m{1.25cm}
                |>{\centering\arraybackslash}m{2.30cm}|}
\hline
            {\textbf{Config.}}     &   {\textbf{Existing inputs}}
        &   {\textbf{New inputs}}   &   {\textbf{Significance of Difference (existing vs. new)}}
        \\
\hline

            {$v$}     & \multicolumn{2}{c|}{0.617 (0.053)} &{N/A}  \\
\hline
            {$v p\textprime$}     &{0.630 (0.037)} &{0.636 (0.055)} &{[-0.062, 0.074]}  \\
\hline
            {$v a\textprime$}     &{0.616 (0.055)} &{0.622 (0.033)} &{[-0.060, 0.072]} \\
\hline
            {$v p\textprime a\textprime$}     &{0.630 (0.038)} & {0.648 (0.027)} &{[-0.030, 0.066]} \\
\hline
\end{tabular}